\documentclass{article} 

\usepackage{iclr2025_conference,times}

\usepackage{amsmath,amsfonts,bm}

\def\eqref#1{equation~\ref{#1}}

\def\1{\bm{1}}

\DeclareMathAlphabet{\mathsfit}{\encodingdefault}{\sfdefault}{m}{sl}
\SetMathAlphabet{\mathsfit}{bold}{\encodingdefault}{\sfdefault}{bx}{n}

\usepackage[numbers,sort&compress]{natbib}
\usepackage{subcaption}
\usepackage{pifont}
\usepackage{wrapfig}
\usepackage{multirow}

\usepackage{makecell}
\usepackage{booktabs}
\usepackage{url}
\usepackage{graphicx}
\usepackage{algorithm}
\usepackage{algorithmic}
\usepackage{amsfonts} 
\usepackage{multirow}
\usepackage{amsmath}
\usepackage{bbding}
\usepackage{pifont}
\usepackage{booktabs}
\usepackage{graphicx}
\usepackage{xcolor}
\usepackage{multirow}
\usepackage{capt-of}
\usepackage{graphicx, amssymb, caption, overpic, textpos}
\usepackage[english]{babel}
\usepackage[pagebackref=true,breaklinks=true,letterpaper=true,colorlinks=blue,citecolor=blue,linkcolor=blue,bookmarks=false]{hyperref}
\usepackage{tikz}
\definecolor{lightred}{rgb}{0.961, 0.874, 0.867} 
\definecolor{lightgreen}{rgb}{0.549, 0.851, 0.616} 

\definecolor{first}{HTML}{547CB1} %
\definecolor{improve}{HTML}{1E73C4} %
\newcommand{\firstone}[1]{\colorbox{first!25}{#1}}
\newcommand{\secondone}[1]{\colorbox{gray!15}{#1}}

\newcommand{\third}[1]{\colorbox{lightred}{#1}}
\newcommand{\fourth}[1]{\colorbox{lightgreen}{#1}}
\hypersetup{
     colorlinks=true,
     linkcolor=red,
     filecolor=blue,
     citecolor=blue,      
     urlcolor=cyan
     }
\title{Accurate Forgetting for All-in-One \\ Image Restoration Model}
\author{Xin Su, Zhuoran Zheng$^{*}$
\\
Fuzhou University, Sun Yat-sen University \\
\texttt{\{suxin4726@gmail.com\}} \\
\texttt{\{zhengzr@njust.edu.cn\}}
}

\iclrfinalcopy 
\begin{document}

\maketitle

\thanks{This article aims to break down the wall between the fields of image restoration \newline and privacy protection. 
Here we introduce a motto to inspire us: \newline \textit{As fruit needs not only sunshine but cold nights and chilling showers to ripen it, \newline so character needs not only joy but trial and difficulty to mellow it.\newline}}

\begin{abstract}
Privacy protection has always been an ongoing topic, especially for AI.
Currently, a low-cost scheme called Machine Unlearning forgets the private data remembered in the model.
Specifically, given a private dataset and a trained neural network, we need to use e.g. pruning, fine-tuning, and gradient ascent to remove the influence of the private dataset on the neural network.
Inspired by this, we try to use this concept to bridge the gap between the fields of image restoration and security, creating a new research idea.
We propose the scene for the All-In-One model (a neural network that restores a wide range of degraded information), where a given dataset such as haze, or rain, is private and needs to be eliminated from the influence of it on the trained model.
%
%
%
%
Notably, we find great challenges in this task to remove the influence of sensitive data while ensuring that the overall model performance remains robust, which is akin to directing a symphony orchestra without specific instruments while keeping the playing soothing.
Here we explore a simple but effective approach: Instance-wise Unlearning through the use of adversarial examples and gradient ascent techniques.
Our approach is a low-cost solution compared to the strategy of retraining the model from scratch, where the gradient ascent trick forgets the specified data and the performance of the adversarial sample maintenance model is robust.
Through extensive experimentation on two popular unified image restoration models, we show that our approach effectively preserves knowledge of remaining data while unlearning a given degradation type.
\end{abstract}

\section{Introduction}
Many organizations leverage user data to train AI models across various fields, from entertainment to healthcare. 
To curb potential abuses, legislation like the European Union’s General Data Protection Regulation (GDPR),  the California Consumer Privacy Act (CCPA), and Canada's Consumer Privacy Protection Act (CPPA) mandates the erasure of user data upon request, safeguarding privacy rights.
However, researchers and customers~\cite{Vadhan2017,10.5555/3361338.3361358} believe that simply removing private data is not enough and that the influence of that data on AI models also needs to be removed.
Usually, this requires retraining the AI model on the remaining dataset, but the cost is exorbitant.
To alleviate this problem of training the model from scratch, Machine Unlearning is proposed, which aims to remove the effect of private data on the model at a low cost based on the given private data and other information.
Recent advances in the field of machine unlearning have introduced various innovative techniques, such as those leveraging the Fisher Information Matrix~\cite{9157084}, NTK theory~\cite{9577384}, and gradient update storage~\cite{yoon2018lifelong}, as well as error-maximizing noise~\cite{chundawat2023zero,tarun2023fast}, teacher-student frameworks~\cite{chundawat2023can,tarun2023deep,kurmanji2024towards}, and parameter attenuation during inference~\cite{chundawat2023zero}. In addition, these methods have also been extended to generative models~\cite{zhang2024unlearncanvas,li2024machine}, including text-to-image diffusion and large language models, through approaches like model editing and layer unlearning.
Indeed, the build-up of these techniques has triggered a new thought, i.e., image restoration tasks also seem to suffer from the need to eliminate the influence of private data on the model.
In this paper, we do our best to explore the issue of privacy protection in the field of image restoration.
%

To facilitate the design of our method, we chose the all-in-one task in the field of image restoration~\cite{ 9879292, yan2023textualpromptguidedimage, yang2023languagedrivenallinoneadverseweather,conde2024instructir} as a case.
The reason for choosing this task is that the all-in-one model can restore multiple degradation scenarios, and we treat one of the degradation types as a private dataset, which is similar to a multi-classification task.
It is worth noting that we also consider eliminating specified datasets in a class rather than all of them, but the results are difficult to evaluate.
Faced with this scenario, we started with using gradient ascent on a given dataset to remove the dehazing or deraining ability of a trained all-in-one model.
Unfortunately, the ability of the all-in-one model to reconstruct other degraded image types is severely compromised.
Although the cost of this approach is low, the model also collapses, for which we try to introduce a small number of other pairs of degenerate-type datasets to smooth the model's performance.
In addition, we introduce adversarial attack samples to improve the robustness of the model, and the results show that the all-in-one model lost the ability to specify the dataset, but the performance of other abilities is improved.
Our experiments are evaluated on two all-in-one models, and our approach effectively degrades the model capabilities using just a single consumer-grade GPU (no more than 2 hours for fine-tuning).
Extensive experiments revealed the limitations of our approach, i.e., the model is virtually ineffective within the first epoch when iterating using the gradient ascent algorithm and changing the learning rate is difficult to work.
In future work, we consider the problem of removing the influence of a specified part of a dataset in a single-class image restoration model (image restoration models that can only address one type of degradation).
\textbf{Our contributions:} 

\textbf{(1)} To the best of our knowledge, we are the first to introduce privacy protection issues in the field of image restoration, to raise awareness of security issues among image restoration researchers.


\textbf{(2)} We introduce an instance-wise unlearning method by increasing the loss for the deleting degradation type of images and their clean images, using only the pre-trained model and datasets within a lower-specification configuration.

\textbf{(3)} We propose a model-agnostic adversarial regularization technique aimed at neutralizing the impact of deleted data. This method employs the targeted use of adversarial examples to ensure that the removal of certain data does not adversely affect the overall restoration capabilities.
\section{Related Work}
\label{sec:background}

\textbf{Machine unlearning.} 
Machine unlearning is proposed by \cite{7163042}, which aims at protecting machine learning models
from extraction attacks, involves the process of removing specific data from a model in a manner that
ensures the data appears as though it were never part of the training set. 
According to the level of forgetting, existing machine unlearning methods can be
categorized into: (1) Exact unlearning~\cite{cao2015towards,bourtoule2021machine}; (2) Approximate unlearning~\cite{nguyen2020variational,chien2022certified}.
Existing works~\cite{ginart2019making,brophy2020dart,mahadevan2021certifiable} focus on approximating unlearning linear/logistic regression, k-means clustering, and random forests. 
The landscape of machine unlearning is diverse, encompassing a range of applications from text and image classification to complex systems like federated learning and recommender models.
Despite the extensive exploration of machine unlearning across diverse domains, including text classification and image-to-image generation, a critical area remains untouched: the end-to-end unlearning in image restoration tasks.
%

\textbf{All-in-One image restoration.}
Image restoration is a fundamental and long-standing problem in computer vision, focusing on recovering a high-quality image from its corrupted counterpart.
Recent studies~\cite{jiang2023autodir,ai2024multimodalpromptperceiverempower,conde2024instructir} seek to use a single framework to handle multiple degradations simultaneously. 
Such methodologies are trained to spot and remedy diverse degradation problems concurrently.
AirNet~\cite{AirNet} utilizes contrastive learning to train an additional degradation encoder, which guides an all-in-one restoration network by identifying the degradation types. 
Similar to AirNet,~\cite{10204770,potlapalli2023promptir,luo2023controlling,Yan2023TextualPG,cui2024adair} both utilize uniquely designed prompts to
guide their networks.
For these all-in-one restoration assistant methods, some training dataset limitations exist in the practical applications of all-in-one models.
Naturally, we would aim to eliminate the restoration capabilities trained on low-quality datasets, ensuring that the all-in-one assistant methods perform optimally with high-quality data.
To achieve this goal, we delve into the concept of machine unlearning.

\textbf{Adversarial examples.}
In light of vulnerabilities in deep learning models~\cite{Pravin_2021}, some studies have proposed a series of adversarial attack methods.
Adversarial examples were initially proposed by~\cite{szegedy2014intriguingpropertiesneuralnetworks}, and the classic method
for generating them, called the Fast Gradient Sign Method (FGSM), proved to be a simple and effective approach. 
Since then, various methods~\cite{madry2019deeplearningmodelsresistant,andriushchenko2020squareattackqueryefficientblackbox,croce2020minimallydistortedadversarialexamples} have been proposed to generate adversarial examples. 
In this paper, we introduce adversarial samples whose purpose is to smooth the performance of the model, and in addition, we try the data augmentation approach whose effect is not significant.
%
%

\section{Method}
\label{sec:method}

\subsection{Overall Pipeline}
\begin{figure}[!t]
    \centering    
    \includegraphics[width=.9\linewidth]{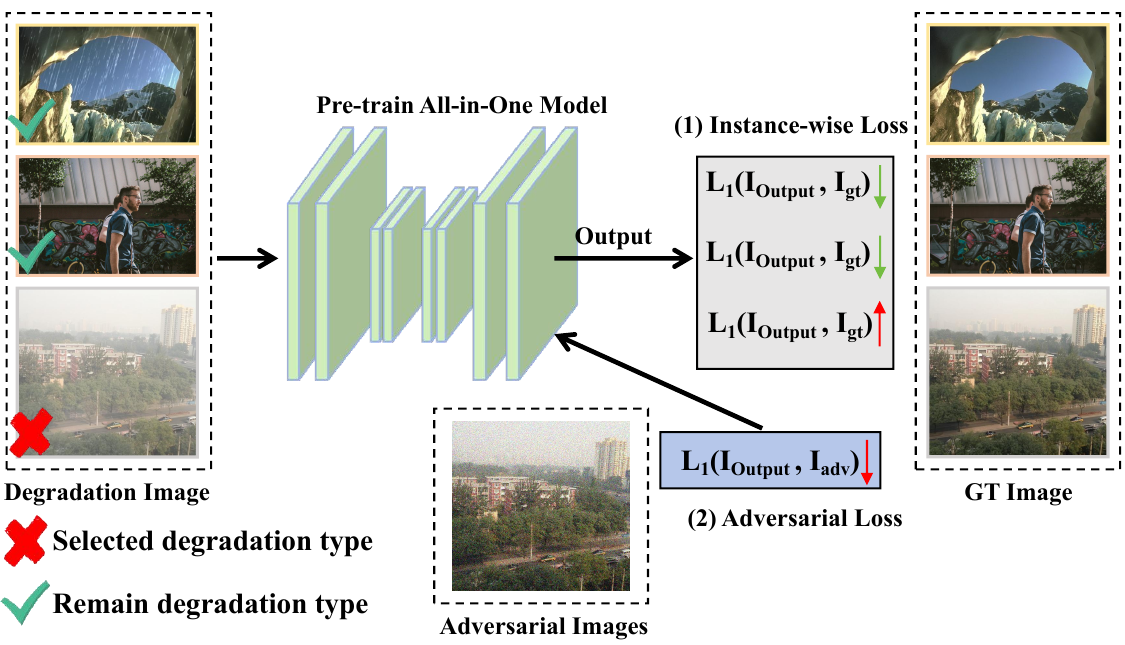}
    \caption{\textbf{Our method.} We start by giving a dataset that needs to be privacy-preserving, other additional degradation information, and an all-in-one neural network that has been pre-trained. Then, we impose an elevated L1 loss on the model's outputs for the targeted degradation type along with their corresponding clean images, thereby encouraging the model to forget the specific degradation pattern during the training process (we employ a gradient-up approach).On the other hand, we introduce adversarial examples and a small number of datasets from other tasks to ensure robust output of the model.}
    \label{fig: framework}
    \vspace{-2.1mm}
\end{figure}

As Fig.~\ref{fig: framework} illustrated, we propose a model-agnostic method for the all-in-one model accurately forgetting specific degradation type data samples used in the model training by incorporating instance-wise unlearning and adversarial examples regularization.

\subsection{Prerequisite} 

Our goal is to ‘unlearn’ specific private datasets by fine-tuning the all-in-one model to eliminate their influence completely.
Here we need two key components: the specified dataset $D$, and the pre-trained neural network $\mathcal{N}$.

\textbf{Dataset and pre-trained model.}
Let \(D=\{x_{i},y_{i}\}_{i=1}^{n}\) be a dataset of various degradation types, where each degradation instance $x_{i}$ is paired with its corresponding clean image $y_{i}$. The index $i \in {1, ..., n}$ represents the total number of degradation types in the dataset.
The all-in-one model $\mathcal{N}$ is trained with the \(D\).
Let \(D_{f}=\{x_{i},y_{i}\}_{i=k} (k \in {1, ..., n})\) be a subset of the \(D\), whose information we want to forget (`unlearn') from the pre-trained all-in-one model $\mathcal{N}$. The remaining data is denoted by \(D_{r}\), whose information is desired to be kept unchanged in the model.
$D_{f}$ and \(D_{r}\) are mutually exclusive and together construct \(D\), i.e., 
\begin{equation}D_{r}\cup D_{f}=D,\end{equation}
\begin{equation}D_r\cap D_f=\varnothing,\end{equation}
where $\varnothing$ denotes the empty set.

\textbf{Adversarial examples.} 
The goal of an adversarial attack on $y_{i}$, which we want to `smooth' the $\mathcal{N}$, is to generate an adversarial example $y'_{i}$.
Then, we optimization (gradient descent method) the $\mathcal{N}$ via the minimize the L1 loss for $y'_{i}$ and $x_{i}$.
For our experiments, we make use of targeted attacks can be described as follows:
\begin{equation}p \sim \mathcal{U}(-0.5,0.5),\end{equation}
\begin{equation}y'_{i}=\mathrm{clamp}(y_{i}+p,\min=0,\max=1),\end{equation}
where $p$ represents a perturbation tensor with elements uniformly distributed between $-0.5$ and $0.5$ and the clamp function ensures that each element of $y_{i}+p$ is within the range $[0, 1]$.

\subsection{Instance-wise unlearning}
Instance-wise unlearning is our core approach, which aims to make pre-trained models $\mathcal{N}$ `repel' the distribution of a given dataset $D_{f}$.
The two goals above can be realized by minimizing the e following loss functions:
\begin{equation}\mathrm{L}_{\mathrm{UL}}({D}_{f};\boldsymbol{\theta})=-\mathrm{L}_{1}(\mathcal{N}(\boldsymbol{x}_{i}),\boldsymbol{y}_{i};\boldsymbol{\theta}),\end{equation}
\begin{equation}\mathrm{L}_{\mathrm{Remain}}(\mathrm{D}_{r};\boldsymbol{\theta})=\mathrm{L}_{\mathrm{1}}(\mathcal{N}(\boldsymbol{x}_{i}),\boldsymbol{y}_{i};\boldsymbol{\theta}),\end{equation}
where $\boldsymbol{\theta}$ denotes the learnable parameters of the model, and $-$ denotes the negative sign whose effect is to make the distribution of such samples $D_{f}$ that the model meets realize forgetting.

\subsection{Regularization using adversarial examples}
The concept of adversarial examples, inspired by the seminal work~\cite{NEURIPS2019_e2c420d9} that introduces perturbations for classification tasks, is adapted in our approach for image restoration.
Here, we specifically tailor the generation of adversarial samples to target and regularize the model's capacity to forget designated degradation types.
In particular, we utilize generated adversarial examples to erase the selected degradation type of restoration capacity via minimizing $L_{1}$ loss function:
\begin{equation}
\mathrm{L}_{\mathrm{UL}}^{\mathrm{Adv}}({D}_f;\boldsymbol{\theta}) = \mathrm{L}_{\mathrm{1}}(\mathcal{N}(\boldsymbol{x}_{i}), \boldsymbol{y'}_{i};\boldsymbol{\theta}),
\end{equation}
where $\boldsymbol{y'}_{i}$ denotes the adversarial examples.

The above procession can be summarized in \textbf{Algorithm}~\ref{alg: Framwork}. The weights of these three minimization functions are discussed in the experimental section. It is worth noting that we tried to use the L2 function to replace L1, but its model performance collapsed.
\begin{algorithm}[!t]
\caption{Our method for accurately forgetting the specific degradation-type restoration ability}
\label{alg: Framwork}
\begin{algorithmic}[1] 
\REQUIRE ~~\\ 
    The pre-train all-in-one model, $\mathcal{N}$;\\
    The training dataset, $D$, which can be divided as two parts, i.e. ${D}_{f}$ and ${D}_{r}$;\\
\ENSURE ~~\\ 
    Accurate forgetting model $\mathcal{N}^{'}$;\\
    Read $\{x_{i},y_{i}\}$ from $D_{f}$ and generate adversarial examples $y'_{i}$;\\
    Read $\{x_{k},y_{k}\}$ from $D_{r}$;\\
    Generate $R_{i} = \mathcal{N}(x_{i})$ and $R_{k} = \mathcal{N}(x_{k})$ during the unlearning procession;\\
    Minimum $-\mathrm{L}_{1}(R_{i},y_{i})+\mathrm{L}_{1}(R_{i},y'_{i})+\mathrm{L}_{1}(R_{k},y_{k})$;\\
\textbf{return} $\mathcal{N}^{'}$; 
\end{algorithmic}
\end{algorithm}

\section{Experiments}
In this section, we present the fundamental setup of our method, the setup of the baseline model, the ablation experiments, and the visualization of the results.

\subsection{Experimental Setup}
\textbf{Datasets and Pre-trained unified model.} 
We evaluate our unlearning method on different restoration tasks: \textit{haze, noise, and rainy}. Table~\ref{tab: datasets statistics} summarises the tasks and the number of training and testing images for each degradation type. We use PromptIR~\cite{potlapalli2023promptir} and AdaIR~\cite{cui2024adair} as the base pre-trained all-in-one model. The compared baselines are as follows: BEFORE corresponds to the pre-trained model before forgetting; OF denotes the ideal model that is retained on $D_{f}$ by our method; OD denotes training datasets is $D$; UT represents the model retrained on $D_{r}$ from scratch.

\begin{wraptable}[8]{r}{6cm}
\caption{\textbf{Datasets Statistics.}}
\vspace{-1mm}
    \label{tab: datasets statistics}
    \centering
    \resizebox{!}{28px}{
        \begin{tabular}{lccc}
            \toprule[1.5pt]
            Dataset & \makecell[c]{Hazy} & \makecell[c]{Noisy} & \makecell[c]{Rainy} \\
            \midrule[1pt]
            \#Train & 3\thinspace4932 & 5\thinspace144 & 2\thinspace4000 \\
            \#Test  & 500 & 68 & 100\\
            \bottomrule[1.5pt]
        \end{tabular}
    }
\end{wraptable}

\textbf{Implementation details.} For unlearning, we use a Adam optimizer ($\beta1$ = $0.9$, $\beta2$ = $0.999$) with learning rate $2e^{-4}$ for \textbf{2} epochs. According to our experimental devices (4 Tesla V100 GPUs), we set a batch size of 16 in the all-in-one setting. During retraining, we still utilize cropped patches of size 128 $\times$ 128 as input, and to augment the training data, random horizontal and vertical flips are applied to the input images.
Our approach can also be run on 4 RTX 3090 GPUs, and our fine-tuning cost is just \textbf{5\%} of training a model from scratch (the cost here is simply the time spent). Following~\cite{Li2023ICLR_uhdfour}, we adopt commonly-used IQA PyTorch Toolbox\footnote{https://github.com/chaofengc/IQA-PyTorch} to compute the \textbf{PSNR}~\cite{PSNR_thu} and \textbf{SSIM}~\cite{SSIM_wang} scores of all compared methods.

\subsection{Main Results}

\textbf{Results on multiple degradation datasets.}
Table~\ref{tab: allinone} shows results before and after forgetting a specific degradation type from PromptIR and AdaIR models pre-trained on Multiple degradation datasets, respectively. As shown in Table~\ref{tab: allinone}, using $D$ as the whole dataset, we achieve a thorough erasure of the influence of the deleting data, while even enhancing the restoration capacity on the remaining degradation types. By setting the $D_{f}$, the restoration capacity of the all-in-one model for all degradation types diminishes. This proves that using our strategy achieves the proposed goal of accurate forgetting. 

\textbf{Results on forgetting effects.} We retrain PromptIR and AadIR on $D_{r}$ from scratch to establish a baseline performance on $D_{f}$. Subsequently, we employ our accurate forgetting strategy to achieve results compared against the test outcomes of a model not trained on the specific degradation type, thereby quantifying the extent of forgetting achieved. 

\begin{figure}[t!] 
    \centering
    \begin{subfigure}[b]{\columnwidth}
        \centering
        \includegraphics[width=\columnwidth]{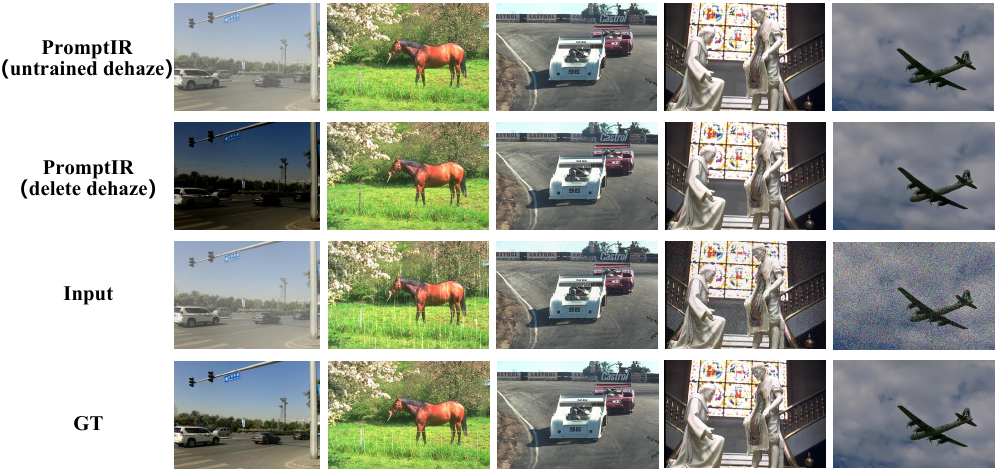}
        \caption{Visual results on PromptIR.}
        \label{fig: promptir_result}
    \end{subfigure}
    \vfill
    \begin{subfigure}[b]{\columnwidth}
        \centering
        \includegraphics[width=\columnwidth]{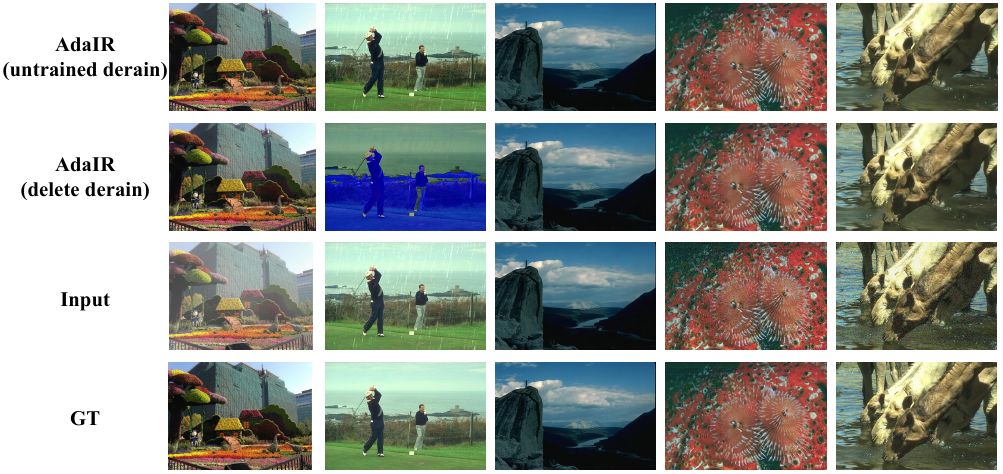}
        \caption{Visual results on AdaIR.}
        \label{fig: adair_result}
    \end{subfigure}
    \caption{Eliminating the dehazing capability from the unified model, the visual outcomes illustrate that our approach can significantly eradicate the impact of deleting data, in contrast to a model that has not been trained on such datasets.}
    \vspace{-4mm}
    \label{fig: unlearining_result}
\end{figure}

As demonstrated in Table~\ref{tab: forgetting effects}, the application of our strategy results in a diminished restoration performance for the specified degradation type, indicating a more thorough forgetting compared to models that have not been trained on this type. Furthermore, our approach leads to an enhancement in the restoration performance for the retained degradation types. This improvement can be attributed to the model's ability to focus its learning capacity on the relevant features after the unlearning process. By eliminating the influence of irrelevant degradation types, the model can allocate more resources to learn and generalize from the pertinent data, thus achieving better performance on the degradation types it is intended to handle. Fig. \ref{fig: unlearining_result} presents visual examples of unlearning based on the PromptIR and AdaIR via our method, which is able to delete the specific restoration capacity while keeping the remaining data performance. 
\begin{table}[!t]
  \centering
  \caption{Comparisons before and after forgetting on PromptIR and AdaIR as a pre-trained model. Our proposed method retains excellent performance on $D_{r}$ while completely forgetting instances in $D_{f}$. The \secondone{column} indicates the original performance of the pre-trained model. The \third{column} represents the deleting degradation type performance while the \fourth{columns} denote the remaining degradation types performance via our method.
}
  \resizebox{\columnwidth}{!}{
  \Huge
  \begin{tabular}{@{}ll|ccccc}
    \toprule[1.5pt]
    &  & Dehazing & Deraining &  \multicolumn{3}{c}{Denoising on BSD68 dataset~\cite{martin2001database_bsd}}  \\
    \vspace{0.5pt}
     & &
  on SOTS~\cite{li2018benchmarking}& on Rain100L~\cite{fan2019general}& $\sigma = 15$ & $\sigma = 25$ & $\sigma = 50$ \\
    \midrule
    \multirow{7}{*}{$\textbf{PromptIR}$} 
    & BEFORE & \secondone{28.89/0.9559} & \secondone{37.34/0.9786} & \secondone{33.98/0.9333} & \secondone{31.32/0.8886} &  \secondone{28.05/0.7769}  \\
    & $\textrm{OF}_\textrm{UL}^\textrm{Dehaze}$ & 5.13~/~0.0337 & 6.43~/~0.0112 &  7.14~/~0.0297 & 7.90~/~0.0701 & 9.75~/~0.1434   \\
    & $\textrm{OF}_\textrm{{UL}}^{\textrm{Derain}}$ & 7.62~/~0.3015 & 6.25~/~0.0030 & 6.71~/0.0094 & 9.59~/~0.8436 &  21.37/0.6665  \\
    & $\textrm{OF}_{\textrm{UL}}^{\textrm{Denoise}}$ & 4.72~/~0.1387 & 5.00~/~0.0981 & 5.11~/~0.0890 & 5.11~/~0.0890 & 5.11~/~0.0890 \\
    & $\textrm{OD}_{\textrm{UL}}^{\textrm{Dehaze}}$ & \third{11.47/0.4647} & \fourth{37.25/0.9714} &  \fourth{33.94/0.9331} & \fourth{31.29/0.8886} & \fourth{28.03/0.7770}   \\
    & $\textrm{OD}_{\textrm{UL}}^{\textrm{Derain}}$ & \fourth{30.09/0.9571} & \third{11.00/0.4547} & \fourth{33.95/0.9329} & \fourth{31.30/0.8883} &  \fourth{28.04/0.7969}  \\
    & $\textrm{OD}_{\textrm{UL}}^{\textrm{Denoise}}$ & \fourth{30.09/0.9569} & \fourth{37.22/0.9776} & \third{11.85/0.6668} & \third{11.66/0.6177} & \third{11.61/0.9865} \\
    \midrule
    \multirow{7}{*}{$\textbf{AdaIR}$} 
    & BEFORE & \secondone{30.90/0.9792} & \secondone{38.02/0.9808} & \secondone{34.01/0.9338} & \secondone{31.34/0.8892} &  \secondone{28.06/0.7978}  \\
    & $\textrm{OF}_{\textrm{UL}}^{\textrm{Dehaze}}$ & 9.03~/~0.3245 & 5.11~/~0.0028 &  28.40/0.9015 & 26.74/0.8322 & 23.72/0.6793   \\
    & $\textrm{OF}_{\textrm{UL}}^{\textrm{Derain}}$ & 20.36/0.7722 & 6.26/0.0030 & 25.14/0.8774 & 27.85/0.8436 &  26.01/0.7230  \\
    & $\textrm{OF}_{\textrm{UL}}^{\textrm{Denoise}}$ & 27.83/0.9712 & 12.35/0.6825 & 4.18~/~0.2589 & 4.17~/~0.2582 & 4.17~/~0.2581 \\
    & $\textrm{OD}_{\textrm{UL}}^{\textrm{Dehaze}}$ & \third{12.40/0.5247} & \fourth{37.23/0.9730} &  \fourth{34.00/0.9335} & \fourth{31.33/0.8886} & \fourth{28.03/0.7970}   \\
    & $\textrm{OD}_{\textrm{UL}}^{\textrm{Derain}}$ & \fourth{34.59/0.9868} & \third{9.81~/~0.4523} & \fourth{33.95/0.9326} & \fourth{31.30/0.8881} &  \fourth{28.03/0.7964}  \\
    & $\textrm{OD}_{\textrm{UL}}^{\textrm{Denoise}}$ & \fourth{33.97/0.9865} & \fourth{37.22/0.9776} & \third{11.85/0.6668} & \third{11.66/0.6177} & \third{11.61/0.9865} \\
    \bottomrule[1.5pt]
  \end{tabular}}
  \label{tab: allinone}
\end{table}

\begin{table}[!t]
  \centering
  \caption{Comparisons retrained on $D_{r}$ from scratch and forgetting based on PromptIR and AdaIR. Our proposed method retains excellent performance on $D_{r}$ while completely forgetting instances in $D_{f}$. The \firstone{column} indicates the untrained on $D_{f}$ of the unified model's performance. The \third{column} represents the deleting degradation type performance while the \fourth{columns} denote the remaining degradation types performance via our method.}
  \resizebox{\columnwidth}{!}{
  \Huge
  \begin{tabular}{@{}ll|ccccc}
    \toprule[1.5pt]
    &  & Dehazing & Deraining &  \multicolumn{3}{c}{Denoising on BSD68 dataset~\cite{martin2001database_bsd}}  \\
    \vspace{0.5pt}
     & &
  on SOTS~\cite{li2018benchmarking}& on Rain100L~\cite{fan2019general}& $\sigma = 15$ & $\sigma = 25$ & $\sigma = 50$ \\
    \midrule
    \multirow{6}{*}{$\textbf{PromptIR}$} 
    & $\textrm{UT}_{\textrm{Dehaze}}$ & \firstone{17.37/0.845} & 39.32/0.986 & 34.26/0.937 & 31.61/0.895 &28.37/0.810    \\
    & $\textrm{UT}_{\textrm{Derain}}$ & 30.09/0.975 & \firstone{24.35/0.975} & 33.69/0.928 & 31.03/0.880 & 27.74/0.777   \\
    & $\textrm{UT}_{\textrm{Denoise}}$ & 30.36/0.973 & 35.12/0.969 & \firstone{20.74/0.519} & \firstone{18.26/0.368} & \firstone{14.19/0.199} \\
    & $\textrm{OD}_{\textrm{UL}}^{\textrm{Dehaze}}$ & \third{11.47/0.465} & \fourth{37.25/0.971} &  \fourth{33.94/0.933} & \fourth{31.29/0.889} & \fourth{28.03/0.777}   \\
    & $\textrm{OD}_{\textrm{UL}}^{\textrm{Derain}}$ & \fourth{30.09/0.957} & \third{11.00/0.455} & \fourth{33.95/0.933} & \fourth{31.30/0.888} &  \fourth{28.04/0.797}  \\
    & $\textrm{OD}_{\textrm{UL}}^{\textrm{Denoise}}$ & \fourth{30.09/0.957} & \fourth{37.22/0.978} & \third{11.85/0.667} & \third{11.66/0.618} & \third{11.61/0.987} \\
    \midrule
    \multirow{6}{*}{$\textbf{AdaIR}$} 
    & $\textrm{UT}_{\textrm{Dehaze}}$ & \firstone{15.92/0.802} & 38.22/0.983 &  34.31/0.938 & 31.67/0.896 &28.42/0.811   \\
    & $\textrm{UT}_{\textrm{Derain}}$ & 30.89/0.980 &  \firstone{24.39/0.795} & 34.11/0.935 & 31.48/0.892 & 28.19/0.802  \\
    & $\textrm{UT}_{\textrm{Denoise}}$ & 30.54/0.978 & 38.44/0.983 & \firstone{20.84/0.543} & \firstone{18.41/0.3822} & \firstone{14.16/0.200} \\
    & $\textrm{OD}_{\textrm{UL}}^{\textrm{Dehaze}}$ & \third{12.40/0.525} & \fourth{37.23/0.973} &  \fourth{34.00/0.934} & \fourth{31.33/0.889} & \fourth{28.03/0.797}   \\
    & $\textrm{OD}_{\textrm{UL}}^{\textrm{Derain}}$ & \fourth{34.59/0.987} & \third{9.81~/~0.452} & \fourth{33.95/0.933} & \fourth{31.30/0.888} &  \fourth{28.03/0.797}  \\
    & $\textrm{OD}_{\textrm{UL}}^{\textrm{Denoise}}$ & \fourth{33.97/0.987} & \fourth{37.22/0.978} & \third{11.85/0.667} & \third{11.66/0.618} & \third{11.61/0.987} \\
    \bottomrule[1.5pt]
  \end{tabular}}
  \label{tab: forgetting effects}
\end{table}
\begin{figure}[t!]
    \centering
    \begin{subfigure}[b]{0.48\columnwidth}
        \centering
        \includegraphics[width=0.9\columnwidth]{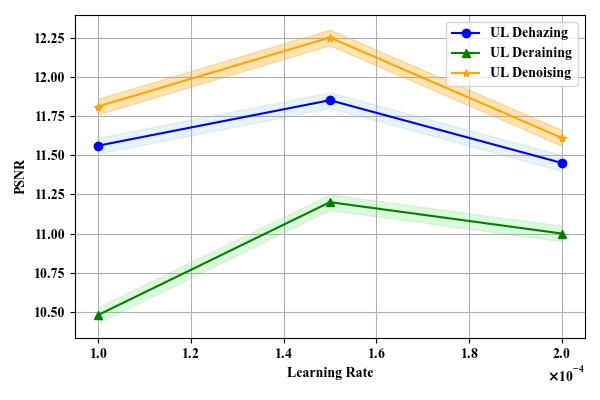}
        \caption{Comparison of unlearning results with different learning rates.}
        \label{fig: lr}
    \end{subfigure}
    \hfill
    \begin{subfigure}[b]{0.48\columnwidth}
        \centering
        \includegraphics[width=0.9\columnwidth]{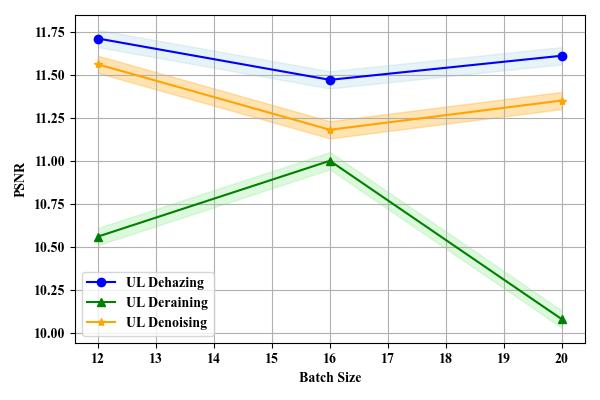}
        \caption{Comparison of unlearning results with different batch size.}
        \label{fig: bs}
    \end{subfigure}
        \vfill
        \begin{subfigure}[b]{\columnwidth}
            \centering
            \resizebox{0.95\columnwidth}{!}{
                \begin{tabular}{@{}ll|ccccc}
        \toprule[1.5pt] 
        \multirow{2}{*}{\textbf{Learning Rate}}&& Dehazing & Deraining &  \multicolumn{3}{c}{Denoising on BSD68 dataset~\cite{martin2001database_bsd}}  \\
        \vspace{0.5pt}
        &&
        on SOTS~\cite{li2018benchmarking}& on Rain100L~\cite{fan2019general}& $\sigma = 15$ &   $\sigma = 25$ & $\sigma = 50$ \\
        \midrule
        \multirow{3}{*}{$\mathbf{1} \times \mathbf{10}^{-4}$}
         &$\textrm{OD}_{\textrm{UL}}^{\textrm{Dehaze}}$& \third{11.56/0.467} & 37.10/0.977 & 33.95/0.933 & 31.30/0.889 & 28.03/0.797 \\
         &$\textrm{OD}_{\textrm{UL}}^{\textrm{Derain}}$& 30.30/0.958 & \third{10.48/0.475} & 33.94/0.933 & 31.30/0.889 & 28.04/0.797 \\
         & $\textrm{OD}_{\textrm{UL}}^{\textrm{Denoise}}$ & 30.14/0.957 & 37.29/0.978 & \third{11.82/0.561} & \third{11.78/0.518} & \third{11.81/0.423} \\
         \multirow{3}{*}{$\mathbf{1.5} \times \mathbf{10}^{-4}$}
         &$\textrm{OD}_{\textrm{UL}}^{\textrm{Dehaze}}$& \third{11.85/0.484} & 37.26/0.977 & 33.96/0.933 & 31.31/0.889 & 28.05/0.798 \\
         &$\textrm{OD}_{\textrm{UL}}^{\textrm{Derain}}$& 30.05/0.958 & \third{11.42/0.459} & 33.94/0.933 & 31.30/0.889 & 28.04/0.797 \\
         & $\textrm{OD}_{\textrm{UL}}^{\textrm{Denoise}}$ & 30.20/0.958 & 37.27/0.978 & \third{12.30/0.694} & \third{12.22/0.643} & \third{12.25/0.546}\\
         \multirow{3}{*}{$\mathbf{2} \times \mathbf{10}^{-4}$}
         &$\textrm{OD}_{\textrm{UL}}^{\textrm{Dehaze}}$& \third{11.47/0.465} & 37.25/0.971 & 31.29/0.889 & 31.31/0.887 & 28.03/0.777 \\
         &$\textrm{OD}_{\textrm{UL}}^{\textrm{Derain}}$& 30.15/0.957 & \third{11.00/0.455} & 33.95/0.931 & 31.30/0.888 & 28.04/0.797 \\
         & $\textrm{OD}_{\textrm{UL}}^{\textrm{Denoise}}$ & 30.09/0.957 & 37.22/0.978 & \third{11.85/0.667} & \third{11.66/0.618} & \third{11.61/0.987}\\
        \bottomrule[1.5pt]
    \end{tabular}
            }
            \caption{Results on different learning rates.}
            \label{tab: lr}
        \end{subfigure}
    \vfill
        \begin{subfigure}[b]{\columnwidth}
            \centering
            \resizebox{0.95\columnwidth}{!}{
                \begin{tabular}{@{}ll|ccccc}
        \toprule[1.5pt] 
        \multirow{2}{*}{\textbf{Batch size}}&& Dehazing & Deraining &  \multicolumn{3}{c}{Denoising on BSD68 dataset~\cite{martin2001database_bsd}}  \\
        \vspace{0.5pt}
        &&
        on SOTS~\cite{li2018benchmarking}& on Rain100L~\cite{fan2019general}& $\sigma = 15$ &   $\sigma = 25$ & $\sigma = 50$ \\
        \midrule
        \multirow{3}{*}{$\mathbf{12}$}
         &$\textrm{OD}_{\textrm{UL}}^{\textrm{Dehaze}}$& \third{11.47/0.465} & 37.25/0.971 & 33.94/0.933 & 31.29/0.889 & 28.03/0.777 \\
         &$\textrm{OD}_{\textrm{UL}}^{\textrm{Derain}}$& 30.09/0.957 & \third{11.00/0.455} & 33.95/0.933 & 31.30/0.888 & 28.04/0.797 \\
         & $\textrm{OD}_{\textrm{UL}}^{\textrm{Denoise}}$ & 30.09/0.957 & 37.22/0.977 & \third{11.85/0.667} & \third{11.66/0.618} & \third{11.61/0.512}\\
         \multirow{9}{*}{}\multirow{3}{*}{$\mathbf{20}$}
         &$\textrm{OD}_{\textrm{UL}}^{\textrm{Dehaze}}$& \third{11.61/0.479} & 37.21/0.977 & 33.95/0.933 & 31.30/0.889 & 28.04/0.797 \\
         &$\textrm{OD}_{\textrm{UL}}^{\textrm{Derain}}$& 29.94/0.956 & \third{10.08/0.366} & 33.92/0.933 & 31.28/0.888 & 28.03/0.796 \\
         & $\textrm{OD}_{\textrm{UL}}^{\textrm{Denoise}}$ & 29.67/0.955 & 37.28/0.978 & \third{11.01/0.531} & \third{11.18/0.531} & \third{11.35/0.423}\\
         \multirow{3}{*}{$\mathbf{16}$}
         &$\textrm{OD}_{\textrm{UL}}^{\textrm{Dehaze}}$& \third{11.47/0.465} & 37.25/0.971 & 31.29/0.889 & 31.31/0.887 & 28.03/0.777 \\
         &$\textrm{OD}_{\textrm{UL}}^{\textrm{Derain}}$& 30.15/0.957 & \third{11.00/0.455} & 33.95/0.931 & 31.30/0.888 & 28.04/0.797 \\
         & $\textrm{OD}_{\textrm{UL}}^{\textrm{Denoise}}$ & 30.09/0.957 & 37.22/0.978 & \third{11.85/0.667} & \third{11.66/0.618} & \third{11.61/0.987}\\
        \bottomrule[1.5pt]
    \end{tabular}
            }
            \caption{Results on different batch size.}
            \label{tab: bs}
        \end{subfigure}
    \caption{The effects of learning rate and batch size on the model's unlearning and restoration performance across various image processing tasks.}
    \vspace{-4mm}
    \label{fig:lr_bs}
\end{figure}

\textbf{Effectiveness of hyper-parameter setting.}
To demonstrate the robustness of our method, we delve into optimizing the batch size, learning rate, and weighting of two loss functions to ensure the effective forgetting of designated data while preserving the model's overall restoration capabilities. Table \ref{tab: lr} and \ref{tab: bs} show, that the robustness of our method against variations in hyperparameters such as learning rate and batch size, demonstrates consistent performance across different settings.

\begin{table*}[ht]
    \centering
    \resizebox{0.96\columnwidth}{!}{
    \small
    \begin{tabular}{@{}lll|ccccc}
        \toprule[1.5pt] 
        \multirow{2}{*}{$\textbf{L}_\textbf{adv}$}&\multirow{2}{*}{$\textbf{L}_\textbf{ins}$}&& Dehazing & Deraining &  \multicolumn{3}{c}{Denoising on BSD68 dataset~\cite{martin2001database_bsd}}  \\
        \vspace{0.5pt}
        &&&
        on SOTS~\cite{li2018benchmarking}& on Rain100L~\cite{fan2019general}& $\sigma = 15$ &   $\sigma = 25$ & $\sigma = 50$ \\
        \midrule
        \multirow{3}{*}{\CheckmarkBold}&\multirow{3}{*}{\XSolidBrush}
         &$\textrm{OD}_{\textrm{UL}}^{\textrm{Dehaze}}$& \third{30.22/0.9580} & 37.48/0.9790 &  33.99/0.9337 & 31.33/0.8891 & 28.06/0.788 \\
         &&$\textrm{OD}_{\textrm{UL}}^{\textrm{Derain}}$& 30.22/0.9583 & \third{37.53/0.9791} & 33.99/0.9336 & 31.33/0.8891 & 28.06/0.7880 \\
         && $\textrm{OD}_{\textrm{UL}}^{\textrm{Denoise}}$ & 29.78/0.9551 & 37.50/0.9792 & \third{33.98/0.9337} & \third{31.33/0.8893} & \third{28.07/0.7984} \\
        \multirow{3}{*}{\XSolidBrush}&\multirow{3}{*}{\CheckmarkBold}
         &$\textrm{OD}_{\textrm{UL}}^{\textrm{Dehaze}}$& \third{14.69/0.6682} & 37.32/0.9778 & 33.96/0.9334 & 31.31/0.8869 & 28.05/0.7976 \\
         &&$\textrm{OD}_{\textrm{UL}}^{\textrm{Derain}}$& 30.15/0.9574 & \third{13.49/0.6484} & 33.94/0.9329 & 31.30/0.8882 & 28.04/0.7964 \\
         && $\textrm{OD}_{\textrm{UL}}^{\textrm{Denoise}}$ & 30.15/0.9574 & 37.37/0.9782 & \third{15.69/0.7983} & \third{15.33/0.7188 }& \third{15.13/0.6304}\\
        \multirow{3}{*}{\CheckmarkBold}&\multirow{3}{*}{\CheckmarkBold} 
         &$\textrm{OD}_{\textrm{UL}}^{\textrm{Dehaze}}$ & \third{11.47/0.4647} & 37.25/0.9714 &  33.94/0.9331 & 31.29/0.8886 & 28.03/0.7770  \\
    && $\textrm{OD}_{\textrm{UL}}^{\textrm{Derain}}$ & 30.09/0.9571 & \third{11.00/0.4547} & 33.95/0.9329 & 31.30/0.8883 &  28.04/0.7969 \\
    && $\textrm{OD}_{\textrm{UL}}^{\textrm{Denoise}}$ & 30.09/0.9569 & 37.22/0.9776 & \third{11.85/0.6668} & \third{11.66/0.6177} & \third{11.61/0.9865} \\
        \bottomrule[1.5pt]
    \end{tabular}}
    \caption{Ablating the each designed regularization approach.}
    \vspace{-4mm}
    \label{tab: ablation}
\end{table*}

\textbf{Effectiveness of regularization with weight.} We propose regularization via the adversarial examples and the instance-wise unlearning. As mentioned above, the initial balance between these two components was set at a 1:1 ratio. However, to further refine our method and explore the impact of different weightings on the model's performance, we have conducted experiments adjusting the ratio of the adversarial examples to instance-wise unlearning. The new ratios tested include 0.5:1 and 1.5:1. As Table \ref{tab: weight} shows, moderately increasing the weight of adversarial examples relative to instance-wise unlearning can effectively balance the forgetting of specific data while maintaining overall model performance.

\begin{table*}[ht]
    \centering
    \resizebox{0.96\columnwidth}{!}{
    \small
    \begin{tabular}{@{}ll|ccccc}
        \toprule[1.5pt] 
        \multirow{2}{*}{$\textrm{W}_\textrm{adv}$ $\colon$ $\textrm{W}_\textrm{ins}$}&& Dehazing & Deraining &  \multicolumn{3}{c}{Denoising on BSD68 dataset~\cite{martin2001database_bsd}}  \\
        \vspace{0.5pt}
        &&
        on SOTS~\cite{li2018benchmarking}& on Rain100L~\cite{fan2019general}& $\sigma = 15$ &   $\sigma = 25$ & $\sigma = 50$ \\
        \midrule
        \multirow{3}{*}{$\textbf{0.5} \colon \textbf{1}$}
         &$\textrm{OD}_{\textrm{UL}}^{\textrm{Dehaze}}$& \third{5.45/0.0067} & 5.18/0.0815 &  6.75/0.0228 & 6.39/0.0228 & 5.32/0.0273 \\
         &$\textrm{OD}_{\textrm{UL}}^{\textrm{Derain}}$& 5.63/0.0881 & \third{6.25/0.0030} & 6.54/0.0820 & 5.56/0.054 & 4.60/0.0272 \\
         & $\textrm{OD}_{\textrm{UL}}^{\textrm{Denoise}}$ & 4.75/0.0223 & 5.05/0.0138 & \third{4.60/0.0179} & \third{4.60/0.0179} & \third{4.60/0.0179} \\
        \multirow{3}{*}{$\textbf{1.5} \colon \textbf{1}$}
         &$\textrm{OD}_{\textrm{UL}}^{\textrm{Dehaze}}$& \third{14.69/0.6682} & 37.32/0.9778 & 33.96/0.9334 & 31.31/0.8869 & 28.05/0.7976 \\
         &$\textrm{OD}_{\textrm{UL}}^{\textrm{Derain}}$& 30.15/0.9574 & \third{13.49/0.6484} & 33.94/0.9329 & 31.30/0.8882 & 28.04/0.7964 \\
         & $\textrm{OD}_{\textrm{UL}}^{\textrm{Denoise}}$ & 30.15/0.9574 & 37.37/0.9782 & \third{15.69/0.7983} & \third{15.33/0.7188} & \third{15.13/0.6304}\\
         \multirow{3}{*}{$\textbf{1} \colon \textbf{1}$}
         &$\textrm{OD}_{\textrm{UL}}^{\textrm{Dehaze}}$& \third{11.47/0.465} & 37.25/0.971 & 31.29/0.889 & 31.31/0.887 & 28.03/0.777 \\
     &$\textrm{OD}_{\textrm{UL}}^{\textrm{Derain}}$& 30.15/0.957 & \third{11.00/0.455} & 33.95/0.931 & 31.30/0.888 & 28.04/0.797 \\
         & $\textrm{OD}_{\textrm{UL}}^{\textrm{Denoise}}$ & 30.09/0.957 & 37.22/0.978 & \third{11.85/0.667} & \third{11.66/0.618} & \third{11.61/0.987}\\
        \bottomrule[1.5pt]
    \end{tabular}}
    \caption{The influence on the ratio of regularization.}
    \vspace{-4mm}
    \label{tab: weight}
\end{table*} 

\section{Ablation Studies} 

We perform ablation studies to show the effectiveness of each designed regularization approach by removing each of them based on PromptIR.
Table~\ref{tab: ablation} indicates that while each regularization approach provides benefits on its own, their combined application maximizes the overall performance, highlighting the synergistic effect of these techniques in enhancing the model's ability to selectively forget and restore information as intended.

\section{Limitations and Conclusion}
Our approach has two unavoided limitations: i) Currently, all-in-one models have several algorithms based on large language models (LLMs) in this paper. Training or fine-tuning these methods requires the inclusion of prompt terms, whereas our strategy has no prompt generation. Therefore, only two all-in-one models are shown in the paper to evaluate the effectiveness of our approach. ii) Our approach requires the small number of remaining datasets $D_{r}$ to be involved in the fine-tuning process. Although we show that using only the specified dataset to participate in fine-tuning is ineffective, it certainly increases the training cost.

In this paper, we present a new research track in the field of image restoration.
To the best of our knowledge, this is the first time that the problem of privacy protection has been introduced in the field of image restoration and a corresponding solution has been proposed.
The purpose of this work is to draw the attention of Low-level researchers to privacy protection.

\bibliography{iclr2025_conference}
\bibliographystyle{iclr2025_conference}


\end{document}